\documentclass{article}

\usepackage{arxiv}

\usepackage[utf8]{inputenc} 
\usepackage[T1]{fontenc}    
\usepackage{hyperref}       
\usepackage{url}            
\usepackage{booktabs}       
\usepackage{amsfonts}       
\usepackage{nicefrac}       
\usepackage{microtype}      
\usepackage{lipsum}
\usepackage{graphicx}
\graphicspath{ {./images/} }

\usepackage{amsmath}
\usepackage{multirow}
\usepackage{subfigure}
\usepackage{arydshln}

\newcommand{\boxx}{{\ttfamily BoXHED} }
\newcommand{\boxxb}{{\ttfamily BoXHED}}

\newcommand{\boxtwo}{{\ttfamily BoXHED2.0} }
\newcommand{\boxtwob}{{\ttfamily BoXHED2.0}}
\newcommand{\boxfuse}{{\ttfamily BoXHEDMM} }
\newcommand{\boxfuseb}{{\ttfamily BoXHEDMM}}

\newcommand{\auroc}{{\ttfamily AUROC} }
\newcommand{\aurocb}{{\ttfamily AUROC}}

\newcommand{\aucpr}{{\ttfamily AUC-PR} }
\newcommand{\aucprb}{{\ttfamily AUC-PR}}

\newcommand{\auct}{{\ttfamily AUCt} }
\newcommand{\auctb}{{\ttfamily AUCt}}

\newcommand{\roc}{{\ttfamily ROC} }

\newcommand{\pr}{{\ttfamily PR} }

\title{Realtime, Multimodal Invasive Ventilation Risk Monitoring using Language Models and \boxxb}

\author{
 Arash Pakbin\textsuperscript{*} \\
  Computer Science \& Engineering\\
  Texas A\&M University\\
  College Station, TX, USA \\
   \And
 Aaron Su\textsuperscript{*} \\
  Computer Science \& Engineering\\
  Texas A\&M University\\
  College Station, TX, USA \\
  \And
 Donald K.K. Lee \\
  Goizueta Business School and \\ Dept. of Biostatistics \& Bioinformatics\\
  Emory University\\
  Atlanta, GA, USA \\  
  \And
 Bobak J. Mortazavi\\
  Computer Science \& Engineering\\
  Texas A\&M University\\
  College Station, TX, USA \\ 
}

\begin{document}
\maketitle
\begin{abstract}
\textbf{ Objective:} realtime monitoring of invasive ventilation (iV) in intensive care units (ICUs) plays a crucial role in ensuring prompt interventions and better patient outcomes. However, conventional methods often overlook valuable insights embedded within clinical notes, relying solely on tabular data. In this study, we propose an innovative approach to enhance iV risk monitoring by incorporating clinical notes into the monitoring pipeline through using language models for text summarization.
\textbf{Results:} We achieve superior performance in all metrics reported by the state-of-the-art in iV risk monitoring, namely: an \auroc of 0.86, an \aucpr of 0.35, and an \auct of up to 0.86. We also demonstrate that our methodology allows for more lead time in flagging iV for certain time buckets.
\textbf{Conclusion:} Our study underscores the potential of integrating clinical notes and language models into realtime iV risk monitoring, paving the way for improved patient care and informed clinical decision-making in ICU settings.
\end{abstract}


\section{INTRODUCTION}

Many ICU patients experience respiratory distress, sometimes requiring iV, which involves intubation to aid breathing \cite{clark2013clinical}. An epidemiological study indicated that from 1993 to 2009, iV usage rose from 179 to 310 cases per 100,000 US adults, correlating with decreased hospital mortality, particularly for patients with Chronic Obstructive Pulmonary Disease and Pneumonia \cite{mehta2015epidemiological}. While critical for survival, prolonged iV can lead to secondary health complications, including ventilator-associated pneumonia \cite{kharel2021ventilator}, and is linked to long-term adverse physical and mental effects, such as increased in-hospital mortality and reduced cognitive function \cite{pham2017mechanical}. Given these risks, iV risk monitoring models are needed to effectively differentiate high-risk from low-risk patients, aiding healthcare professionals in predicting iV needs, enabling early interventions, and determining optimal times for extubation.

\begin{figure*}[t!]
\centering    
	\includegraphics[scale=0.32]{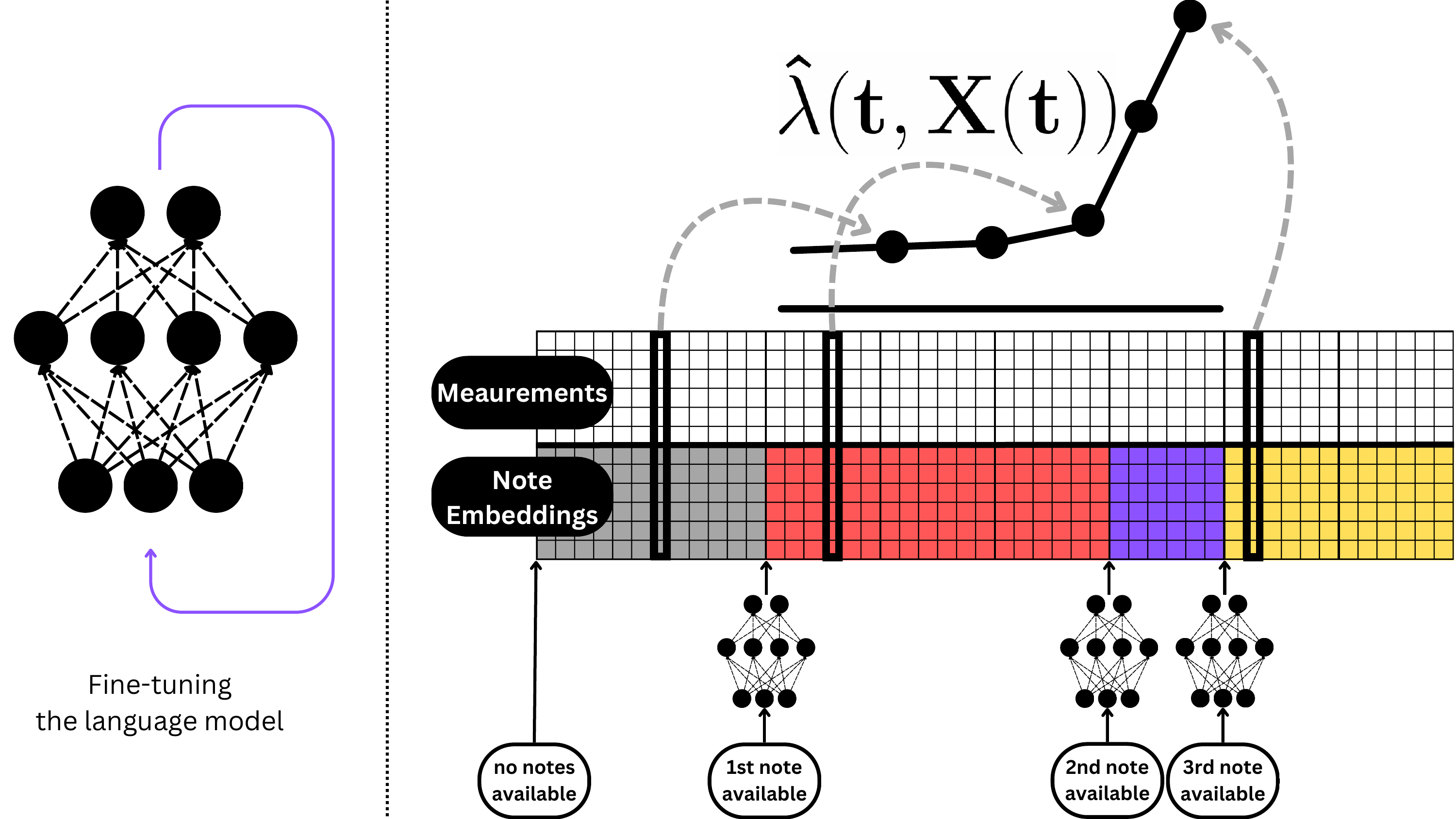}
	\caption{The overall structure of \boxfuseb. It integrates tabular time-series measurements with information extracted from the most recent clinical note. The combined information feed into our realtime risk monitoring system.}    
 \label{fig:overall_structure}
\end{figure*}

Survival analysis is well-suited for estimating the risk of respiratory distress and the time to its occurrence. Most survival models focus on single events \cite{lee2019dynamic, salerno2022high}, while iV can recur during an ICU stay. Traditional models often rely on static time data \cite{cox1972regression}, missing the potential of evolving time-series patient data. The method \boxtwo \cite{pakbin2021boxhed, LCI} is a fully nonparametric approach designed to handle recurrent events and time-dependent covariates, making it ideal for iV complexities and enabling realtime applications. We previously presented a conference work on a realtime iV monitoring system using \boxtwob \cite{pakbin2023predicting}, achieving state-of-the-art performance. However, \boxtwob's reliance on tabular time-series data limits it to demographics and time-varying vital signs, while Electronic Health Records (EHR) provide a richer data source.

Tabular data represent only a portion of the data contained within EHRs; EHR data is comprised of many unstructured data fields as well, including text and images \cite{kong2019managing}. Such unstructured data are rich, and models restricted to using only unstructured data demonstrate high performance in clinical risk prediction tasks \cite{mahbub2022unstructured}. Clinical notes, for example, contain latent medical insights that inform clinical decision making. The introduction of note embeddings through deep learning-based natural language processing methods, has enabled clinical models to leverage the wealth of information contained within notes without resorting to simplistic heuristics or manual feature engineering \cite{susnjak2024applying}. In this study, we advance our preliminary approach \cite{pakbin2023predicting} by extending \boxtwo to incorporate clinical notes in the realtime prognostication of iV risk. Our contributions are as follows:
\begin{itemize}
    \item We fine-tune and deploy Clinical-T5\cite{lehman2023clinical}, a language model pretrained on datasets from MIMIC III\cite{johnson2016mimic} and MIMIC IV\cite{Johnson2021-si} to generate embeddings of notes throughout an admission to the ICU.
    \item We extend \boxtwo to handle note embeddings for multimodal analysis.
    \item We demonstrate a realtime implementation of risk monitoring for the potentially recurrent iV event.  
\end{itemize}  
This study treats clinical notes similarly to structured data, integrating their timing and availability with structured time-series information. The outcome is \boxfuseb, \textbf{\boxx} \textbf{M}ulti \textbf{M}odal, a system that delivers realtime estimates of the time to an iV event by continuously incorporating new multimodal data as it becomes available. Fig.~\ref{fig:overall_structure} shows the overall structure of \boxfuseb. Our findings demonstrate an enhancement in realtime risk monitoring performance across all metrics reported by \cite{pakbin2023predicting}: \aurocb, \aucprb, \auctb, and lead time in flagging patients. 

\subsection{RELATED WORKS}


\textbf{\boxtwob} is a scalable tree-boosted hazard estimator \cite{pakbin2021boxhed} designed to handle time-varying data in survival analysis, in a continuous-time setting. It accommodates complex censoring mechanisms, including recurrent events.
\boxtwo models the hazard function with the tree ensemble
\begin{equation}\label{eq:cox}
\hat\lambda(t,X(t)) = \exp\left\{F_0 + \nu\sum_{m=0}^{M-1} g_{m}(t,X(t))\right\}
\end{equation}
where $F_0$ is the best constant fit for the log-hazard, and each $g_{m}(t,x)$ is a tree learner that is built sequentially in a boosting fashion. \boxtwo supports multi-core CPU/GPU computing, enabling it to be utilized for realtime applications, including the one in this work.

\textbf{Recent Approaches to iV.}
Recent efforts in iV risk monitoring typically frame the problem as a single-time event, predicting whether it will occur during the ICU stay. For example, \cite{bastidas2024performance} predicts iV as a binary task using data from the first 6 hours, while \cite{schwager2024machine} uses the first 24 hours. Although \cite{zhang2024intelligent} employs a rolling window approach to account for the recurrent nature of iV prediction, their models treat different time windows independently, missing temporal dependencies. Our preliminary work \cite{pakbin2023predicting} developed a model for iV risk estimation using \boxtwo on MIMIC-IV data \cite{johnson2016mimic}, overcoming these limitations but relying solely on structured time-series measurements and under-utilizing EHR data richness.

\textbf{Language models in Medicine}
Clinical-T5 \cite{lehman2023clinical} is a variant of the T5 (Text-To-Text Transfer Transformer) model \cite{raffel2020exploring}, pretrained on clinical text from MIMIC-III and MIMIC-IV. It is designed for healthcare applications, supporting summarization, question answering, and text generation from clinical narratives. T5, a standard encoder-decoder Transformer with a maximum sequence length of 512, is pretrained on heuristically cleaned web text. We employ this model to encode notes as time-series for \boxtwob.

\section{RESULTS}

This section discusses cohort selection, focusing on time-series data and clinical note extraction in Section~\ref{sec:cohort}. We then evaluate \boxfuseb's performance in realtime iV risk monitoring by comparing various note embedding settings to find the optimal dimensionality, using \auroc and \aucpr as key metrics. After identifying the best model, we compare it with the state-of-the-art approach from \cite{pakbin2023predicting}, highlighting the benefits of integrating clinical notes into risk monitoring. This comparison uses metrics from \cite{pakbin2023predicting}, including \aurocb, \aucprb, and \auctb. Finally, we assess \boxfuseb's realtime utility by examining lead times for accurately flagging true iV occurrences, which is crucial for clinical decision support, and conclude with a discussion on the importance of variables in each model, providing insights into their contributions to risk prediction.

\subsection{Cohort Selection Process}\label{sec:cohort}

The MIMIC-IV dataset \cite{Johnson2021-si} is a public resource with de-identified health data for over 40,000 patients in critical care at Beth Israel Deaconess Medical Center. It includes clinical data such as time-series measurements (e.g., vital signs), lab results, imaging reports, and clinical notes. We apply the MIMIC-III preprocessing pipeline \cite{harutyunyan2019multitask} to extract demographic and vital measurements, along with respiratory indicators relevant to iV, like oxygen flow rates. Categorical variables are one-hot encoded, and missing values are filled by carrying forward the last observation. Risk monitoring starts 24 hours into care, tracking iV recurrence through the cumulative count and time since the last occurrence. The timing of iV acts as labels, with patients receiving surgical tracheotomy censored. We partition the cohort into training (24,764 patients, 30,716 ICU stays) and testing (4,344 patients, 5,352 ICU stays) sets, with 69.0\% of training stays lacking iV events, 28.9\% having one, and 2.1\% having two or more. This preparation follows \cite{pakbin2023predicting} for consistent comparisons.

The MIMIC-IV dataset includes radiology and discharge notes, with an average of 3.7 radiology notes and 1.0 discharge note per ICU stay. Discharge notes summarize hospitalization details but are only available at the end, making them unsuitable for prognostication. In contrast, radiology notes document imaging modalities (e.g., x-rays, CT, MRI, ultrasound) throughout the ICU stay, so we focus on them as our primary clinical notes.

For results comparable to \cite{pakbin2023predicting}, we identified notes in the MIMIC-IV dataset corresponding to our training and test cohorts. For each time-series entry, we matched the ICU stay ID to locate the relevant notes and selected the most recent note before the timestamp, saving its ID. We then checked subsequent entries to see if an iV event occurred after the note. If an event was detected, the note was labeled as 1; otherwise, it was labeled as 0. These note-label pairs will be used to fine-tune our language model.

\begin{table*}[t!]
	\caption{A comparison of performance for different note embedding dimensions in terms of \auroc and \aucpr  (with 95\% C.I.). Larger values are better for both metrics.}	
	\centering
		\begin{tabular}{c|c:ccccc}
			\multirow{2}{*}{\shortstack{Note embed \\dim}} & \multirow{2}{*}{baseline} & \multirow{2}{*}{2} & \multirow{2}{*}{4} & \multirow{2}{*}{8} & \multirow{2}{*}{16} & \multirow{2}{*}{32} \\
            & \\
			\hline
            \auroc & 0.50 & \textbf{0.86} & \textbf{0.86} & \textbf{0.86} & \textbf{0.86} & \textbf{0.86}\\
            \aucpr & 0.07 & \textbf{0.35} & 0.33 & 0.33 & 0.32 & 0.34\\
            \hline
			\multirow{2}{*}{\shortstack{\boxtwo hyperparams\\ (max depth, \# trees)}} & \multirow{2}{*}{--} & \multirow{2}{*}{(3, 125)} & \multirow{2}{*}{(3, 100)} & \multirow{2}{*}{(3, 75)} & \multirow{2}{*}{(3, 100)} & \multirow{2}{*}{(3, 100)} \\
            & \\
		\end{tabular}
	\vskip -0.1in
	\label{tab:auroc_aucpr_note_dims}
\end{table*}

\begin{figure*}[t!]
    \centering 
    \subfigure[\roc plot]
    {\label{fig:roc}\includegraphics[width=0.49\textwidth]{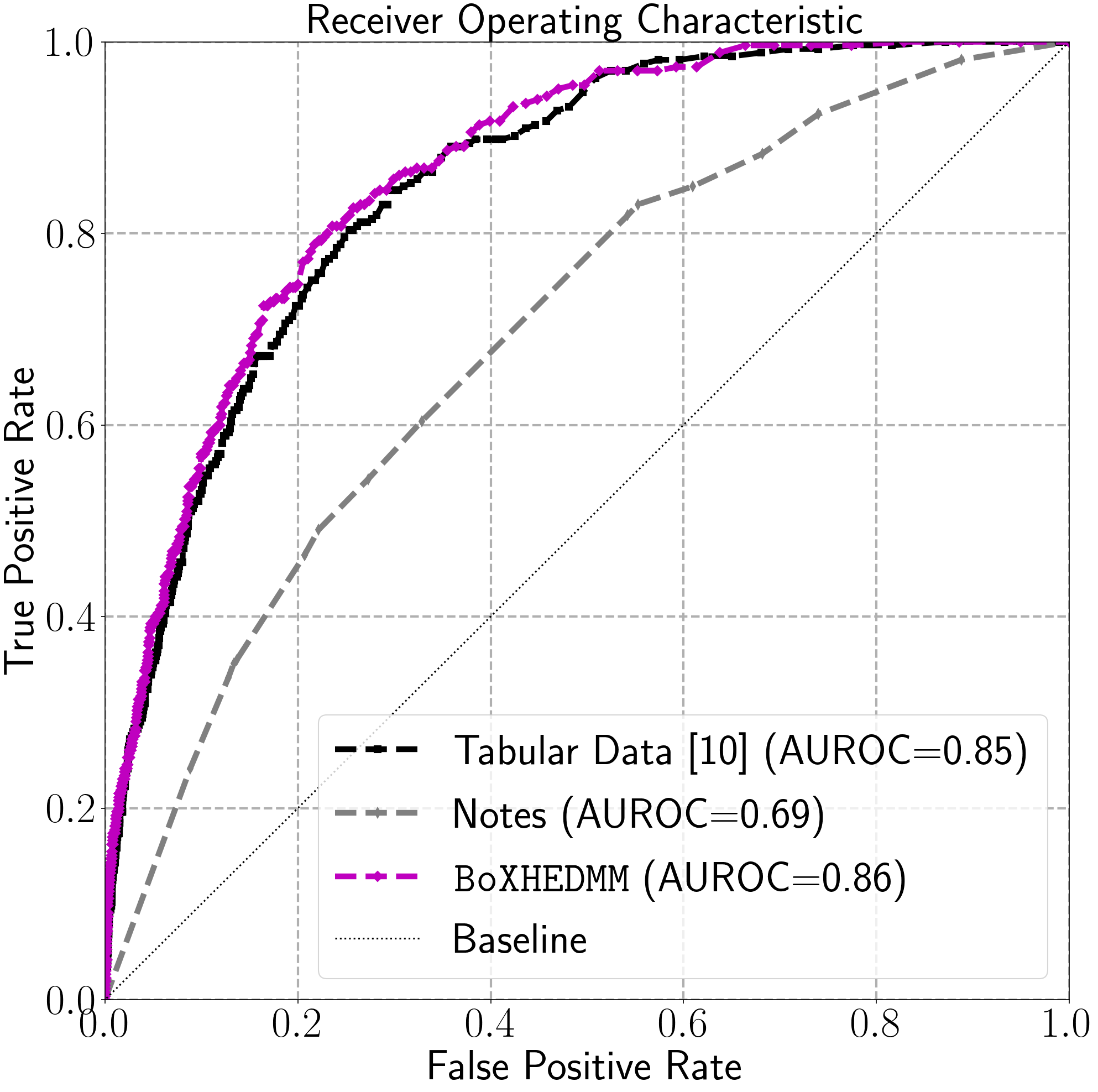}}
    \subfigure[\pr plot]
    {\label{fig:pr}\includegraphics[width=0.49\textwidth]{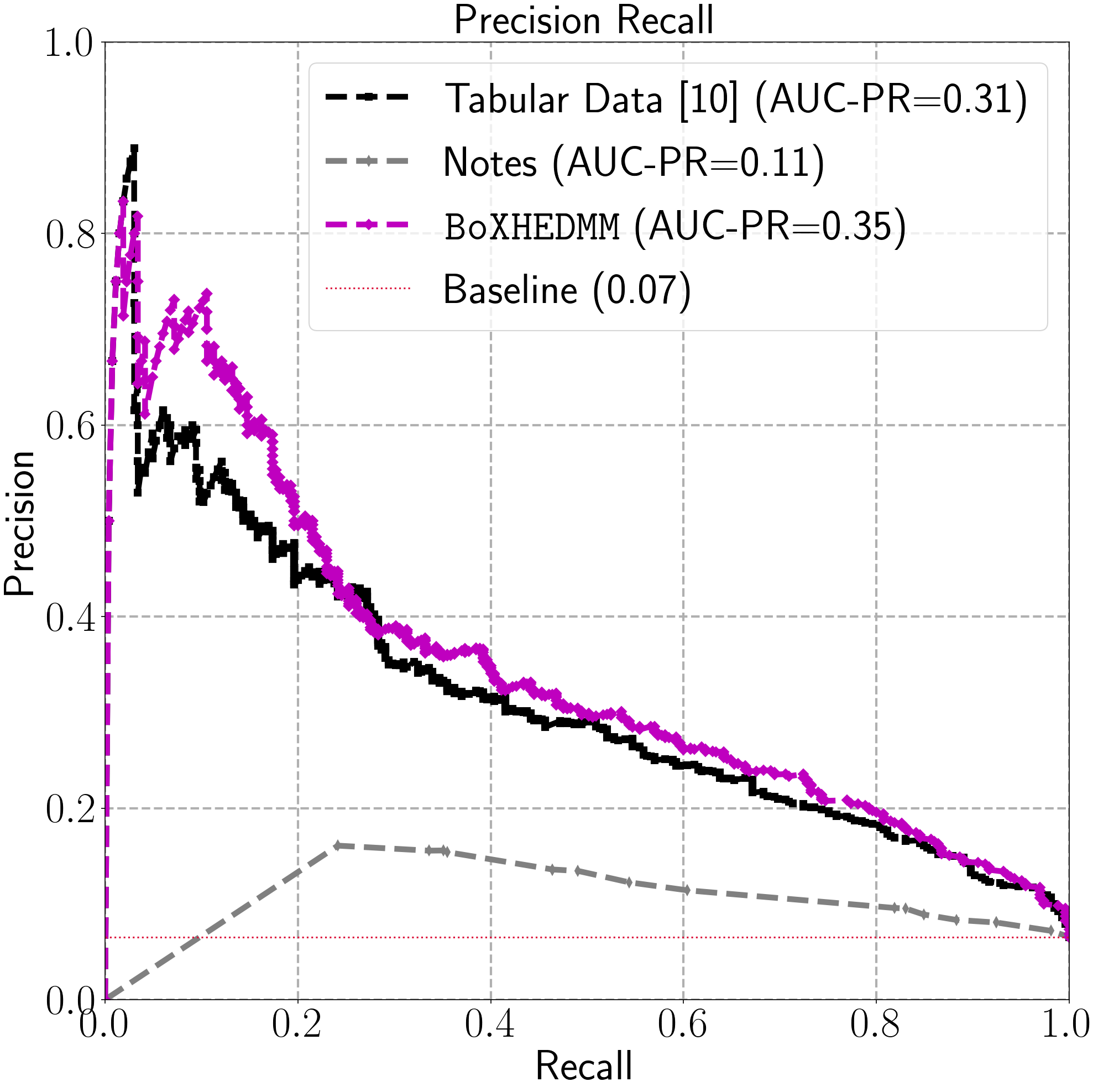}}
    \caption{\roc and \pr plots comparing \boxfuse with the approach from \cite{pakbin2023predicting} for iV prognostication. \auroc and \aucpr values are also provided, where higher values indicate better performance for both metrics.}
    \label{fig:roc_pr}
\end{figure*}

\subsection{Finding the Best Note Embedding Dimensionality}
Table \ref{tab:auroc_aucpr_note_dims} compares \auroc and \aucpr metrics for realtime iV prediction in the MIMIC-IV ICU dataset across note embedding dimensions from 2 to 32. The baseline \auroc is $0.50$, and \aucpr is $0.07$. An embedding dimension of 2 achieves an \auroc of $0.86$ and an \aucpr of $0.35$, outperforming other dimensions. Therefore, we adopt 2 as the default embedding dimension for further analyses, naming the corresponding realtime iV monitoring system \boxfuseb.

\subsection{Realtime iV Risk Monitoring: Comparing against State-of-the-art}
Fig.~\ref{fig:roc_pr} compares the \roc and \pr curves for predicting iV in the MIMIC-IV ICU dataset, including models based on tabular data from \cite{pakbin2023predicting}, clinical notes, and \boxfuseb. The \boxfuse model achieves an \auroc of $0.86$, exceeding the $0.85$ reported by \cite{pakbin2023predicting}. The notes-only model shows a drop in \auroc to $0.69$. For \aucprb, \boxfuse achieves $0.35$, surpassing the $0.31$ from \cite{pakbin2023predicting}, while the notes-only model attains $0.11$.

\begin{figure*}[h]
    \centering
    \begin{minipage}[t]{0.49\textwidth}
        \centering
        \includegraphics[width=\linewidth]{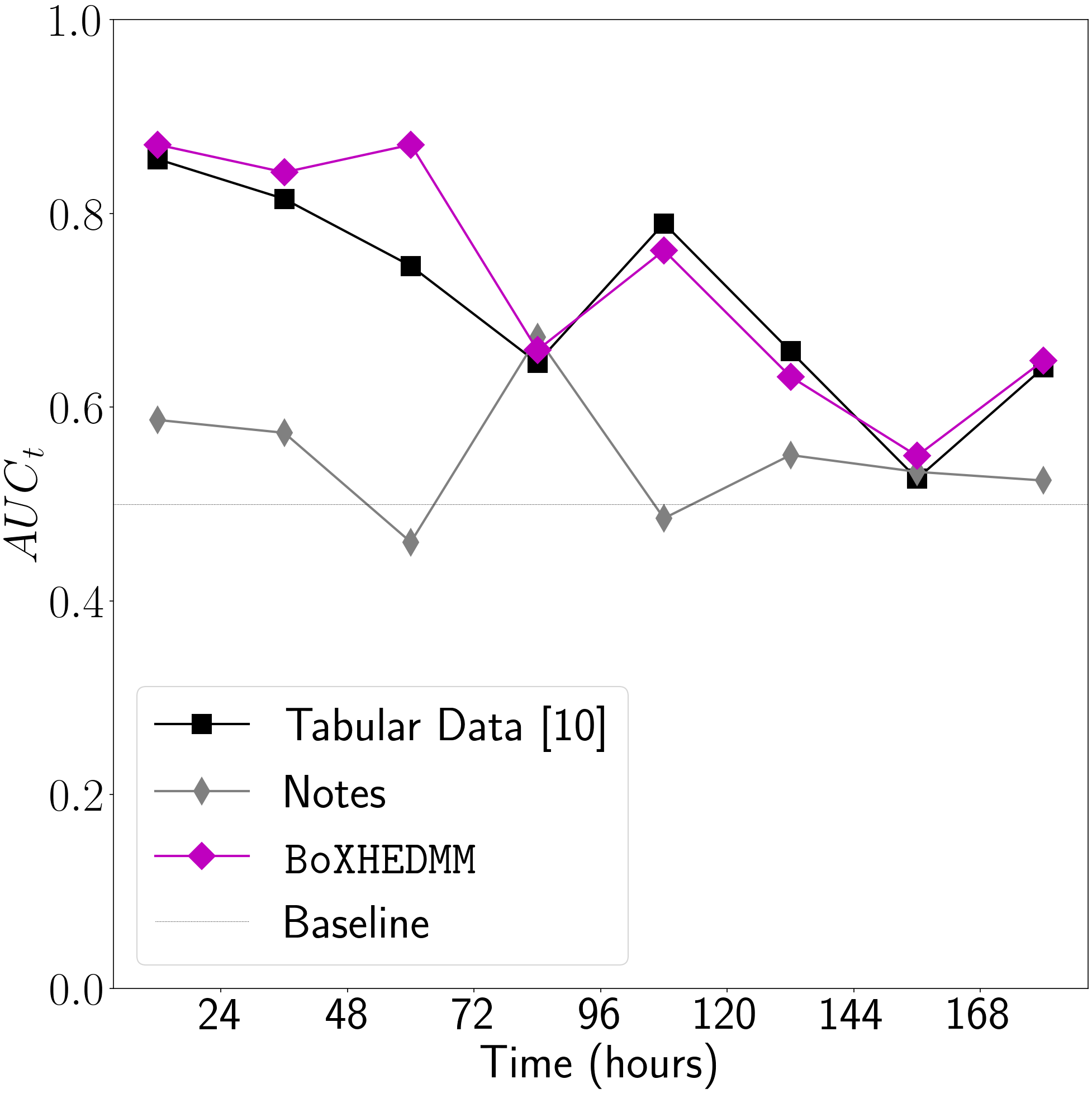}
	\caption{\auct values with 95\% confidence intervals for iV prognostication. A higher \auct value indicates better concordance.}    
    \label{fig:auct}
    \end{minipage}
    \hfill
    \begin{minipage}[t]{0.49\textwidth}  
        \centering
        \includegraphics[width=\linewidth]{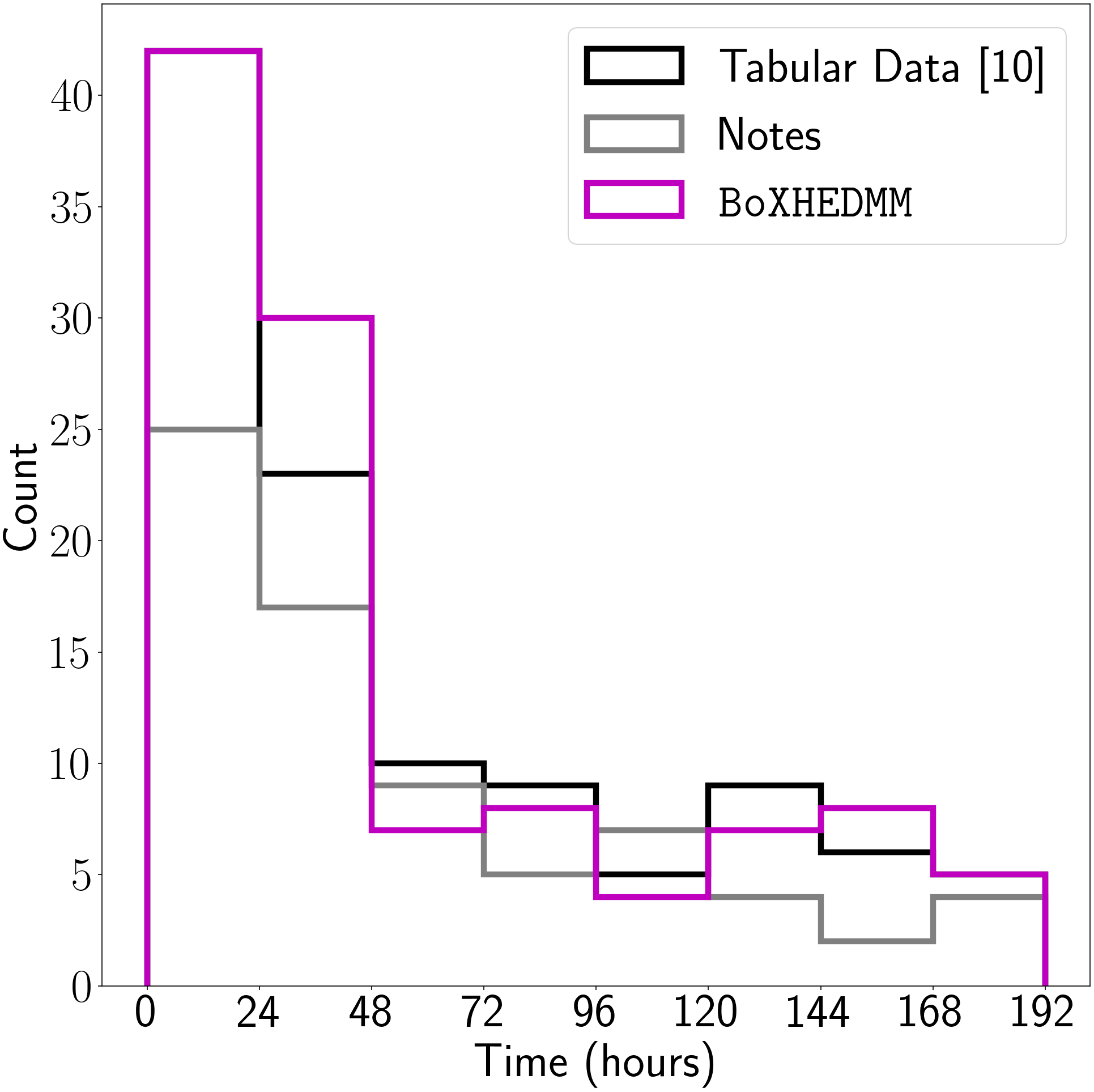}
	\caption{Histogram of lead times for flagging the predicted cases of iV.} 
 \label{fig:lead_time}
    \end{minipage}
\end{figure*}

Across nearly all time intervals, \boxfuse outperforms the method in \cite{pakbin2023predicting} (see Fig.~\ref{fig:auct}), with the most significant improvement in the $24$- to $48$-hour period. The notes-only model consistently underperforms compared to the other two methods regarding \auctb, by a substantial margin.

Fig.~\ref{fig:lead_time} compares lead times (in hours) for accurately flagging iV cases. \boxfuse and the method in \cite{pakbin2023predicting} perform identically in the first 24 hours, while the notes-only model underperforms. In the 24- to 48-hour window, \boxfuse shows a clear advantage over the other two models, but results for longer periods are mixed.

\begin{figure*}[t!]
    \centering 
    \subfigure[Methodology from \cite{pakbin2023predicting}.]{\label{fig:var_imp_base}\includegraphics[width=0.49\textwidth]{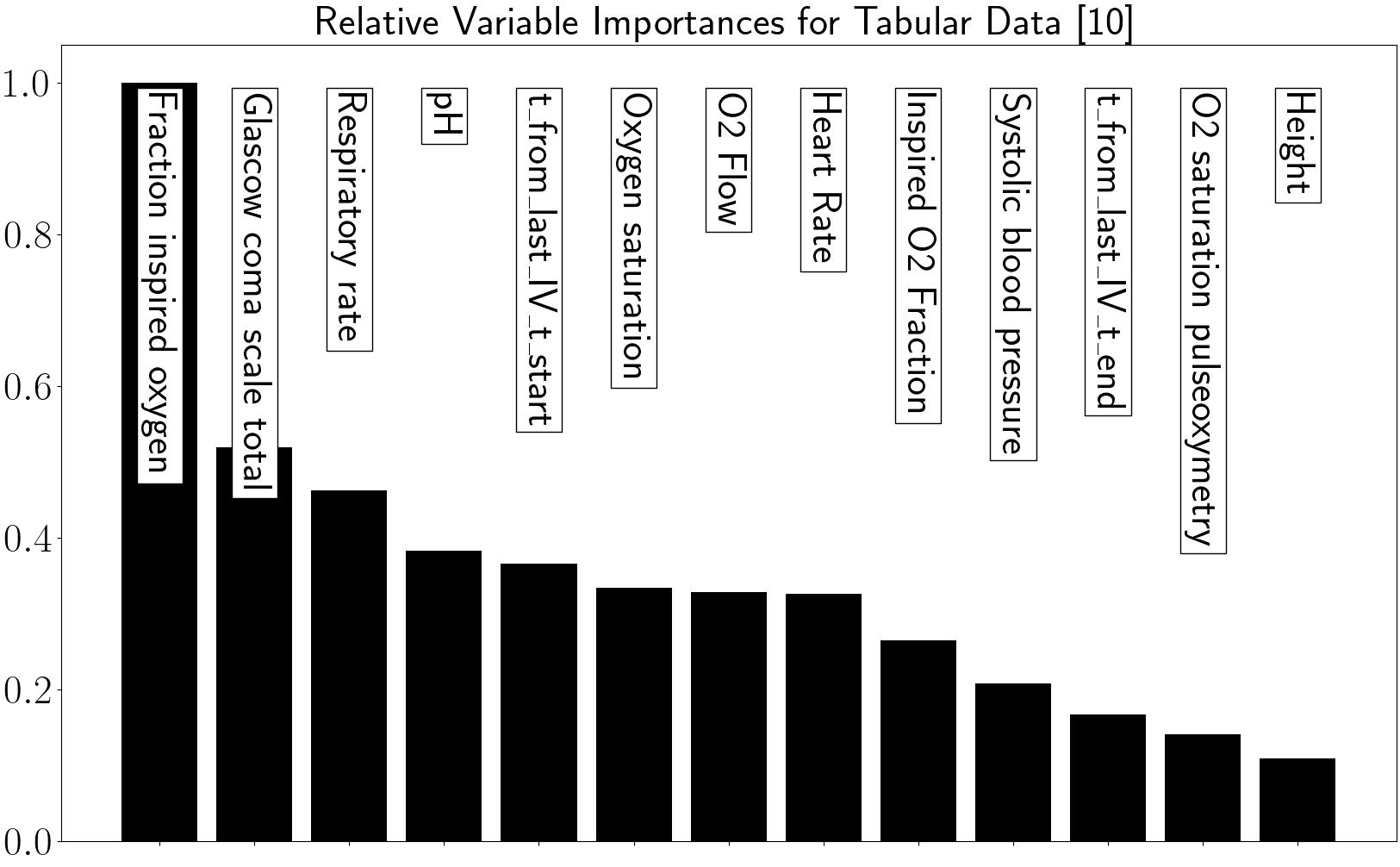}}
    \subfigure[\boxfuse]{\label{fig:var_imp_boxfuse}\includegraphics[width=0.49\textwidth]{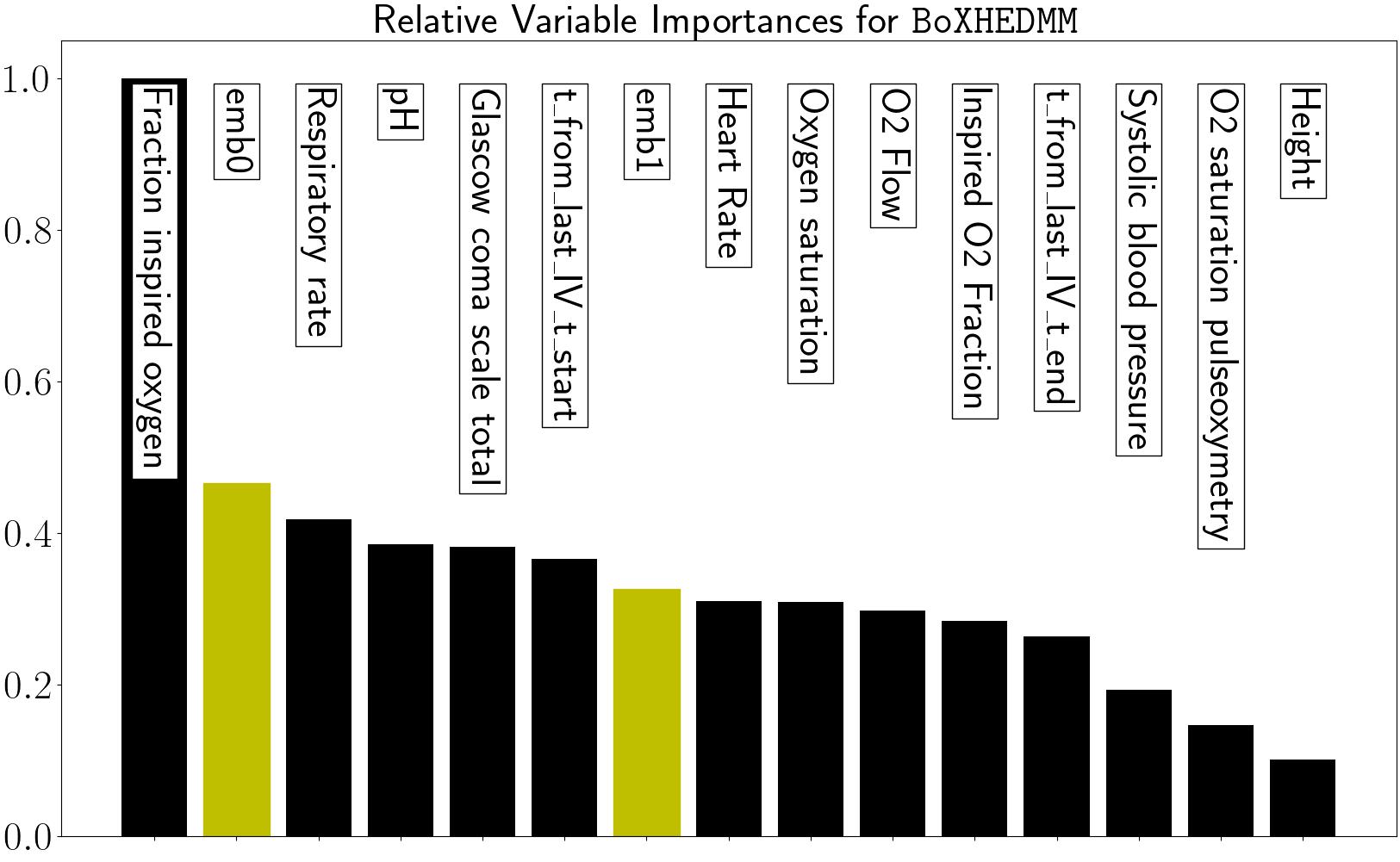}}
    \caption{Comparison of relative variable importances for \boxfuse and the methodology from \cite{pakbin2023predicting}. The plot includes only those variables with relative importances greater than $0.1$. Highlighted bars correspond to notes embeddings.}
    \label{fig:var_imps}
\end{figure*}

Fig.~\ref{fig:var_imps} compares the most relevant variables with a relative importance of at least $0.1$. \boxtwo evaluates variable importance by measuring log-likelihood improvement from splits on each variable during training, normalizing by maximum importance. As shown in Fig.~\ref{fig:var_imp_base}, \cite{pakbin2023predicting} identifies \textit{Fraction of inspired oxygen}, \textit{Glasgow Coma Scale total}, and \textit{Respiratory rate} as key numerical variables, while \boxfuse highlights \textit{Fraction of inspired oxygen}, the first note embedding dimension 'emb0', and \textit{Respiratory rate}.

\section{DISCUSSION}

The 2-dimensional embedding outperforms higher-dimensional embeddings, as shown in Table \ref{tab:auroc_aucpr_note_dims}. While larger embeddings expand the network's parameter space and enhance complexity capture, they did not improve iV risk prediction. Notably, our cross-validation set a maximum tree depth of 4, which was never selected, suggesting that higher-order interactions (beyond order 3) are absent in data from larger embeddings. Thus, the 2-dimensional embedding provides a more compact and effective representation for this task.

The notes-only model consistently underperforms compared to the tabular data-only model \cite{pakbin2023predicting} across all evaluated metrics. This indicates that tabular data provides significantly more relevant information for iV risk monitoring than clinical notes.

\boxfuseb, which integrates tabular data and clinical notes, shows only a marginal improvement over the tabular data-only model \cite{pakbin2023predicting}. This suggests that the notes provide overlapping information with the tabular inputs, limiting the enhancement from their integration.

This study highlights the potential of clinical notes in survival analysis but has limitations. The 512-token limit of Clinical-T5 results in unused note data. Future work could chunk notes for evaluation or utilize models that hanle longer sequences. Exploring various fine-tuning objectives, classifier architectures, and optimization techniques may also enhance performance.

Another limitation is the reliance on the MIMIC-IV dataset, which contains only radiology and discharge notes. In contrast, MIMIC-III includes a broader range of notes, potentially providing richer data. Replicating the study with MIMIC-III could yield more frequent note embeddings but complicate comparisons with \cite{pakbin2023predicting}. As MIMIC-IV evolves, it may incorporate additional note types, enhancing the methodology's future applicability.

\section{CONCLUSION}
Our investigation highlights the advantages of integrating clinical notes with time-series data for realtime iV risk monitoring. By leveraging language models to summarize information from these notes, we introduce a valuable enhancement to the monitoring framework. Utilizing \boxtwob, a leading method in risk estimation, we demonstrate how the rich information within clinical notes can be harnessed by risk estimators. This integration improves the accuracy of realtime iV risk assessment, thereby potentially advancing the quality of patient care.

\section{MATERIALS AND METHODS}
\boxfuse consists of two main components: data fusion and risk estimation. The data fusion component enhances time-series measurements by extracting relevant numerical information from the most recent clinical notes using a language model, integrating it with the time-series data. This dynamic dataset is processed by our realtime risk monitoring system, \boxtwob, to generate continuous risk assessments that incorporate both clinical notes and numerical measurements, providing ongoing risk estimates over time. The overall structure of \boxfuse is shown in \ref{fig:overall_structure}.

This section provides a detailed description of \boxfuseb. We explain the process of extracting numerical information from the clinical notes in Section~\ref{sec:note_embed}. We then describe realtime risk tracking using the fusion of time-series and note-derived information in Section~\ref{sec:boxhed_on_fuse}. Finally, we outline the evaluation methods for the resulting risk assessments in Section~\ref{sec:eval}.


\subsection{Numerically Representing Notes using Language Models}\label{sec:note_embed}

To work with clinical notes, we must first convert the text into numerical vectors in $\mathbb{R}^n$, where $n$ is the encoding's dimensionality. This is done using the encoder from the transformer architecture \cite{vaswani2017attention}. We focused on encoder-only models like BERT \cite{devlin2018bert} or the encoder of models like Clinical-T5 \cite{lehman2023clinical}, prioritizing pretrained language models to save time and computational resources, avoiding extensive pretraining on MIMIC-IV data.

We chose Clinical-T5 for its availability as a pretrained model on MIMIC-IV data, specifically using the Clinical-T5-Base version. This model, initialized from T5-Base and further trained on MIMIC-III and MIMIC-IV, offered a good balance of performance and resource efficiency. Despite larger T5 models, we selected Clinical-T5-Base with 220 million parameters, using only the encoder for note embeddings, which contains 110 million parameters.

Clinical-T5-Base processes sequences of 512 tokens, each embedded in $\mathbb{R}^{768}$, and outputs sequences with the same dimensions. Following BERT \cite{devlin2018bert}, we use the first token's output as the representation for the entire input note, resulting in a note represented as a vector in $\mathbb{R}^{768}$.

We add a classification head to the encoder for two key reasons: to generate a scalar output in $\mathbb{R}$ for fine-tuning and to reduce the 768-dimensional note embeddings to a more manageable size. The hidden layers in the classification head, being smaller than 768, create a more compact note representation.

The classification head takes a 768-dimensional input and outputs a single logit. It has two hidden layers: the second, which serves as our note embedding, has a size $n$ (experimented with), and the first has a size of $\lceil{\sqrt{768*n}}\rceil$ to reduce dimensionality in two roughly equal stages. Each layer is followed by ReLU and dropout of 0.1 to improve robustness.

\begin{figure}[t!]
\centering    
	\includegraphics[scale=0.4]{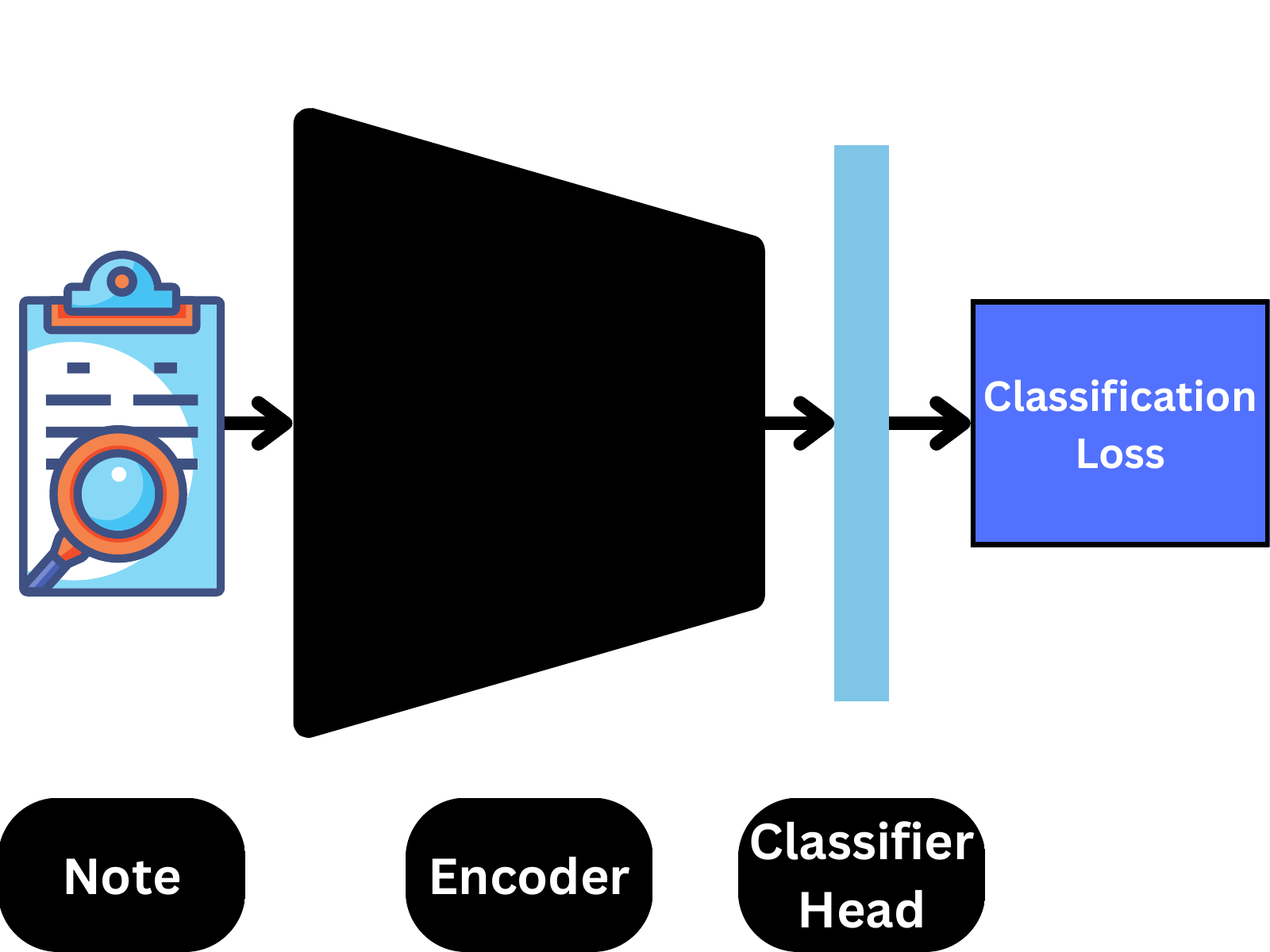}
	\caption{Fine-tuning involves feeding notes into the encoder to generate embeddings, which are processed by the classifier head. Both components are trained using classification loss.}    
 \label{fig:fine-tuning}
\end{figure}

Fine-tuning language models improves task performance, especially by enhancing domain-specific understanding and reducing data and resource needs compared to training from scratch. Here, we fine-tune the encoder to generate note embeddings.

Selecting an efficient fine-tuning objective is key for iV risk monitoring. Due to the complexity of iV as a recurrent event, addressing it fully would require survival analysis, reserved for later stages. Instead, we approximate the problem as a binary classification task: a note is labeled positive if the patient experiences iV at any future point, regardless of timing; otherwise, it's labeled negative. This simplifies the task for fine-tuning, using binary cross-entropy with logits as the loss function.

Since the encoder is pretrained and the classification head is not, we fine-tune in two steps. First, we freeze the encoder and train only the classifier, allowing it to adapt without altering the encoder. After selecting the best classifier checkpoint, we unfreeze the encoder and jointly train both components, retaining the pretrained encoder's benefits while optimizing the model for the task.

We truncate input notes to 512 tokens before fine-tuning. Using an 80/20 train/validation split, we experiment with embedding dimensions $n \in {2, 4, 8, 16, 32}$. Fine-tuning is done on 8 NVIDIA GeForce GTX 1080 Ti GPUs: 8 epochs for the classifier head with a fixed encoder, followed by 16 epochs of joint training. With a batch size of 64 (2 samples per device, 4 gradient accumulation steps), we use the AdamW optimizer (learning rate 0.001, weight decay 0.1) with a cosine schedule and 8 cycles. Checkpoints are saved each epoch, with the best selected by validation loss.

After fine-tuning, we process the most recent note at each timestamp through the fine-tuned Clinical-T5 encoder and classification head to obtain an $n$-dimensional note embedding from the classifier's last hidden layer. These embeddings are integrated with the tabular data based on their timestamps, assigning NaN values if no notes are available. This integration enables the model to incorporate the latest textual information from clinical notes into the time-series analysis, improving the accuracy of risk predictions.

\subsection{Tracking Realtime Risk on Fused Data}\label{sec:boxhed_on_fuse}
For each \( n \in \{0, 2, 4, 8, 16, 32\} \), we create datasets combining time-series measurements with embeddings from recent notes, where \( n=0 \) indicates no note embeddings. We apply \boxtwo to each case and perform 5-fold cross-validation on the training set using a parameter grid for maximum tree depths ranging from 1 to 4 and the number of trees varying from 25 to 500 in increments of 25. The best model is selected based on log-likelihood values using the \textit{one-standard-error rule} (see \S 7.10 of \cite{hastie2009elements}), ensuring robust model selection while accounting for complexity. After identifying optimal hyperparameters, we train \boxtwo on the entire training dataset to estimate the hazard value \(\hat{\lambda}(t, X(t))\), which reflects the realtime risk of iV over time, continuously updating based on the latest data.

\subsection{Evaluating Realtime Risk}\label{sec:eval}

Evaluating time-varying risk is challenging. One approach is to use the methodology from \cite{pakbin2023predicting} to identify high-risk ICU patients whose risk exceeds a certain threshold. The risk monitoring system's performance is evaluated by checking if these flagged patients experience iV. By framing this as a classification task, we utilize metrics such as \aurocb, \aucprb, and \auct \cite{blanche2019c} to assess the system's effectiveness. A more detailed description of the evaluation process follows.

The estimated hazard $\hat\lambda(t,X(t))$ of iV varies during a patient's ICU stay. When it exceeds a threshold $\rho$, the patient is flagged as high-risk, and monitoring stops. If the patient later experiences iV, it counts as a true positive, regardless of the time elapsed. If flagged but never experiences iV, it is a false positive. If the estimated risk remains below $\rho$ and iV occurs, it is a false negative. Lastly, if $\hat\lambda(t,X(t)) < \rho$ throughout and iV does not occur, it is a true negative. This classification based on the threshold $\rho$ allows evaluation of the risk monitoring system's performance.

The Receiver Operating Characteristic (ROC) curve and Precision-Recall Curve are created by varying the threshold $\rho$ across possible risk values. Given the class imbalance, where only 7\% of episodes result in an iV, the Area Under the Precision-Recall Curve (\aucprb) is more informative than \auroc \cite{saito2015precision}. The \aucpr emphasizes true positive identification within the minority class, offering better insight into the model's effectiveness in this imbalanced context.

To measure outcome concordance, we use \auctb, the time-dependent {\ttfamily AUC} defined as:
\[
\text{\ttfamily{AUCt}}(t) = \mathbb{P}\left(\hat{S}_i(t) < \hat{S}_j(t) \middle| \Delta_i = 1, T_i < t \le T_j\right).
\]
Here, $\hat{S}_i(t)$ is the conditional survival probability based on the covariate trajectory ${X_i(u)}{u\in[0,t]}$, and $T_i$ is the event time. The empirical \auct at time $t$ is calculated using pairs $i$ and $j$ such that $T_i < t \le T_j$ and $\Delta_i = 1$, measuring the proportion of pairs where $\hat{S}_i(t) < \hat{S}_j(t)$. \auct is evaluated only at observed event times, and we smooth results by binning and averaging the \auct values.

Following \cite{pakbin2023predicting}, we evaluate the effectiveness of realtime iV risk monitoring by measuring how early patients are flagged in true positive cases. This is done by calculating the time between being flagged as high-risk and experiencing iV, using the $\rho$ value that maximizes the F1-score.

\section*{Acknowledgments}
This work was supported in part by the National Institutes of Health grants R21-EB028486 and R01-HL164405.

\bibliographystyle{unsrt}  
\bibliography{main}  






\end{document}